\documentclass[final]{cvpr}

\pdfoutput=1
\usepackage{times}
\usepackage{epsfig}
\usepackage{graphicx}
\usepackage{amsmath}
\usepackage{amssymb}
\usepackage{booktabs}
\usepackage{microtype}
\usepackage{multirow}
\usepackage{subcaption}
\usepackage[skip=0.5\baselineskip]{caption}

\usepackage{footmisc}

\usepackage[pagebackref=true,breaklinks=true,colorlinks,bookmarks=false]{hyperref}

\def\cvprPaperID{****} 
\def\confYear{CVPR 2021}

\begin{document}

\title{Discerning Generic Event Boundaries in Long-Form Wild Videos}

\author{Ayush K. Rai,
Tarun Krishna,

Julia Dietlmeier,

Kevin McGuinness \\
Alan F. Smeaton,

Noel E. O'Connor
\vspace{10pt}
\\
{Insight Centre for Data Analytics, Dublin City University, Ireland}\\
{\{\texttt{ayush.rai3, tarun.krishna2}\}}\texttt{@mail.dcu.ie}, \\
{\{\texttt{julia.dietlmeier}\}}\texttt{@insight-centre.org} \\
{\{\texttt{kevin.mcguinness, alan.smeaton, noel.oconnor}\}}\texttt{@dcu.ie}\\

}

\maketitle

\begin{abstract}
   Detecting generic, taxonomy-free event boundaries in videos represents a major stride forward towards holistic video understanding. In this paper we present a technique for generic event boundary detection based on a two stream inflated 3D convolutions
   architecture, which can learn spatio-temporal features from videos. Our work is inspired from the Generic Event Boundary Detection Challenge (part of CVPR 2021 Long Form Video Understanding- LOVEU Workshop). Throughout the paper we provide an in-depth analysis of the experiments performed along with an interpretation of the results obtained. The code for this work can be found at \href{https://github.com/rayush7/GEBD}{https://github.com/rayush7/GEBD}
\end{abstract}

\section{Introduction}
It is a natural tendency of humans to perceive  videos as a composition of events like  making breakfast, attending class, watching a movie etc. These events could further be segmented into a sequence of shorter temporal units as studied in cognitive psychology~\cite{tversky2013event}.
Event boundaries comprise of instances of greater change in action, cases indicating completeness of specific goals and sub-goals, occasions where predictability collapses etc. Event boundary detection has a plethora of significant applications in complex action recognition, video summarisation, video editing and ad-cue points
detection for YouTube videos. In the last few years tremendous advancements have been made in action anticipation ~\cite{Miech_2019_CVPR_Workshops, abu2018will}, temporal action detection ~\cite{Chao_2018_CVPR, gao2017cascaded}, segmentation ~\cite{kuehne2014language,10.1007/978-3-319-46487-9_3} and parsing ~\cite{Pirsiavash_2014_CVPR, Shao_2020_CVPR}. However only limited progress has been made when it comes to detecting event boundaries in long form videos due to unavailability of proper task definition and annotations.

In this direction, the generic event boundary detection challenge was organised in the Long Form Video Understanding Workshop at CVPR 2021 to further investigate this task. The challenge uses the newly proposed Kinetics-GEBD~\cite{shou2021generic}, which contains the largest number of boundaries (around 32x  ActivityNet, 8x  EPIC-Kitchens- 100). The boundaries have open vocabulary, contain generic event changes, are in the wild and adhere to human perception diversity. The challenge aims at predicting the timestamps where an event boundary is most likely to occur. The train, validation and test dataset each contained nearly 20,000 videos of duration 10 secs. Each video was annotated by 5 annotators separately for event boundaries and every annotator was given a F1-consistency score (an indicator of annotator rating) as explained in \cite{shou2021generic}. The evaluation protocol used is the relative distance (Rel.Dis) whereas the official metric for the challenge is F1@5\%, which is defined as the F1 score computed with 5\% threshold. Rel.Dis is the error between detected and ground-truth timestamps, divided by the length of the whole video.

\section{Related Work}
Video understanding encapsulates various tasks like action recognition, action anticipation, action detection, temporal action detection, video summarisation, event boundary detection, etc. Shot Boundary Detection ~\cite{10.1016/j.cviu.2009.03.011,souvcek2019transnet} is a long-standing problem to detect shot transitions (zooming in/out, fading in/out effect, camera shot change) in videos which are added during video editing. Although in this task shot boundaries have very well defined vocabulary making them significantly easier than generic event boundary detection.

Temporal Action Detection involves the task of detecting the start and end of action instances in an untrimmed, long video. There have been many standard datasets including THUMOS~\cite{idrees2017thumos}, ActivityNet ~\cite{caba2015activitynet} to address this problem. However all of these have predefined action classes and a fixed norm to define the beginning and end of actions. Some of the works in this direction include ~\cite{Chao_2018_CVPR, gao2017cascaded}. Temporal Action Segmentation refers to the task of labelling the instances of actions in every frame of the video. Some well known benchmarks for this task are GTEA~\cite{lei2018temporal} and 50Salads~\cite{stein2013combining}. Recently a lot of progress has been made in this field ~\cite{kuehne2014language,10.1007/978-3-319-46487-9_3}. Temporal Action Parsing~\cite{Shao_2020_CVPR} focuses on identifying temporal partitions for decomposing an action into sub-actions e.g. parsing an instance of triple jump into segments - run-up, three jumps, and then a reset.

\begin{figure*}
\centering
\includegraphics[width=\textwidth]{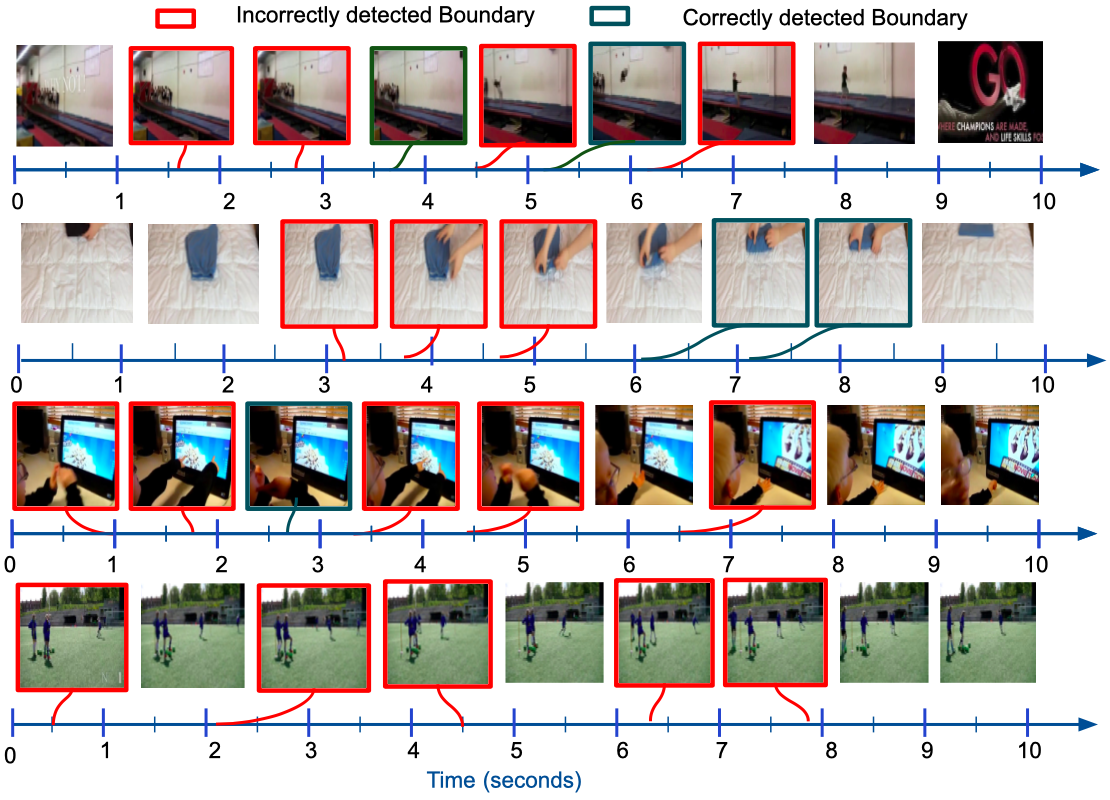}
\caption{Event Boundary detection for a video in gymnastics tumbling, folding clothes, using computer and shooting goal (soccer) classes (top to bottom) in KineticsGEBD (val set) }
\label{fig:results_boundary_detection}
\end{figure*}

\begin{figure*}[t]
\begin{subfigure}{.5\textwidth}
  \centering
  \includegraphics[width=8cm]{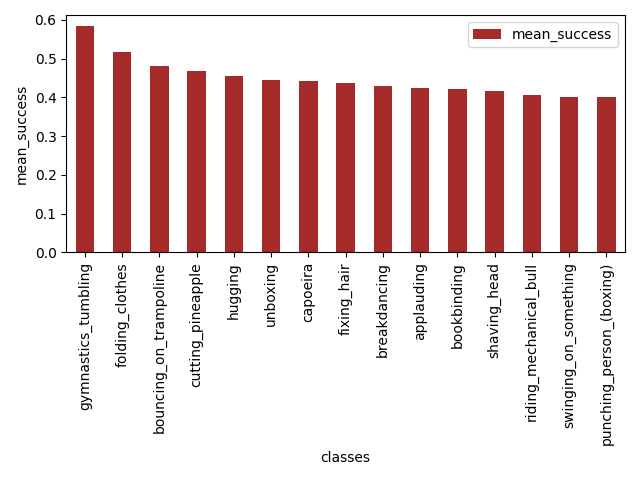}
  \caption{highest mean success}
  \label{fig:sfiga}
\end{subfigure}%
\begin{subfigure}{.5\textwidth}
  \centering
  \includegraphics[width=8cm]{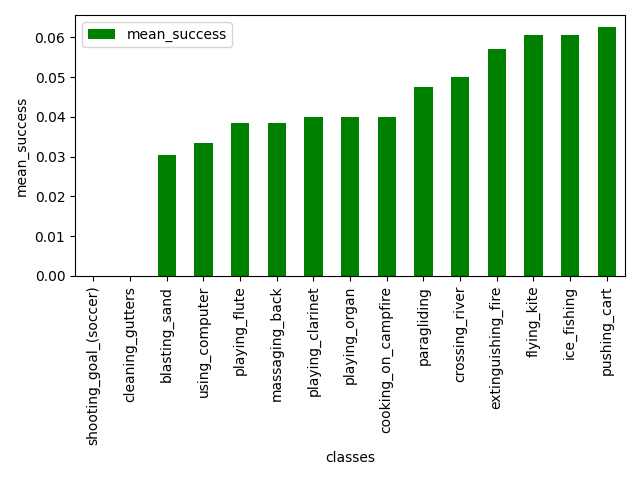}
  \caption{lowest mean success}
  \label{fig:sfigb}
\end{subfigure}
\caption{Top 10 classes in KineticsGEBD (Val set) with highest mean (Fig\ref{fig:sfiga}) and lowest mean (Fig\ref{fig:sfigb}) success for correctly detected boundary.}
\label{fig:success_failure}
\end{figure*}

\begin{table}[ht]
   \caption{Results on Validation Set}
   \label{tab:result_val}
   \resizebox{\columnwidth}{!}{
     \begin{tabular}{@{}lccc@{}}
     \toprule
     Model & F1 Score@5\% & Precision@5\% & Recall@5\% \\ 
     \midrule
     \textit{Baseline \footref{baseline} \cite{shou2021generic}} &  52.11 & 60.44 & 45.81 \\
     \midrule
     \textit{RGB} & 51.39 & 48.20 & 55.04 \\
     \midrule
     \textit{RGB+Flow} & 51.26 & 47.96 & 55.06 \\
     \midrule
     \textit{RGB}$_{Fixed}$ & 50.72 & 42.44 & 63.03 \\
     \bottomrule
 \end{tabular}
 }
 \end{table}

\begin{table}[ht]
   \caption{Results on Test Set}
   \label{tab:result_test}
   \resizebox{\columnwidth}{!}{
     \begin{tabular}{@{}lccc@{}}
     \toprule
     Model & F1 Score@5\% & Precision@5\% & Recall@5\% \\
     \midrule
     \textit{Baseline \footref{baseline} \cite{shou2021generic}} &  58.03 & 69.35 & 49.89 \\
     \midrule
     \textit{\textbf{RGB}} & \textbf{66.05} & \textbf{58.89} & \textbf{75.20} \\
     \midrule
     \textit{RGB+Flow} & 65.95 & 58.58 & 75.45 \\
     \midrule
     \textit{RGB}$_{Fixed}$ & 63.23 & - & - \\
     \bottomrule
 \end{tabular}
 }
 \end{table}

\section{Method}

The objective here is to localise moments into short temporal segments from a long video sequence, where the boundaries for such short segments are often triggered by changes in background, activity, persons, etc., i.e. it has different level of details in both space and time. Thus it is difficult to envision what will be the best approach to tackle such diverse scenarios for boundary detection.

We hypothesise that motion could play a crucial role in detecting such changes. To this end, we decided to exploit I3D ~\cite{carreira2017quo} model, which seemed to be computationally efficient and has been trained on \textit{Kinetics 400} and made publicly available. I3D is a two stream network which relies on 3D ConvNet to learn about temporal patterns from \textit{RGB} stream and further, to boost performance it also exploits an optical flow stream to incorporate motion cues. We use I3D in two settings:

\begin{itemize}
    \item \textit{Fine-tuned I3D}: In this, we considered (1) only \textit{RGB} (2) both \textit{RGB} $+$ \textit{Flow} streams. We used the pre-trained I3D and fine-tuned it for binary classification. 
    \item \textit{Fixed feature extractor}:  Under this regime, we use I3D as a fixed feature extractor, i.e. we consider the output from the penultimate layer before last the \texttt{Conv3d} layer and further augment the network with two non-linear layers \texttt{I3D->ReLU(1024)->ReLU(256)->2}. This was done for \textit{RGB} stream only.   
\end{itemize}
 
 To calculate the optical flow we used an \textit{OpenCV} ~\cite{opencv_library} based \textit{Farneback} algorithm for computing optical-flow. The flow for all the images was computed offline and was fed along with corresponding \textit{RGB} images in a two-stream framework. The input to the RGB and Flow streams is   $\mathbf{x} \in \mathbb{R}^{(B \times 2m \times 3 \times 224 \times 224)}$\footnote{Input format as followed in PyTorch} and $\mathbf{y} \in \mathbb{R}^{(B \times 2m \times 2 \times 224 \times 224)}$ respectively, where $B$ stands for batch size and $m=5$; i.e. we consider \textit{m} frames before and after time-stamp \textit{t} in order to classify whether the frame at time-stamp \textit{t} is a boundary or background.  

\textbf{Baseline:} We considered a ResNet50 backbone trained on ImageNet which was fine-tuned with  \textit{Pairwise Boundary classifier} (PC) ~\cite{shou2021generic} as baseline\footnote{\label{baseline}Our baseline results were lower compared to ~\cite{shou2021generic}, which was probably because we only had availability to 17,159 training, 15,176 validation and 17,254 test videos.}. The input to PC is a concatenation of two vectors which are an average of feature representation of \textit{m} frames before and after time-stamp \textit{t}. 

\textbf{Reproducibility:} We trained our model on 2 NVIDIA GeForce RTX 2080Ti GPUs for 16 epochs and batch size of 16 using the Adam~\cite{kingma2014adam} optimiser. The learning rate was chosen to be 0.0001 which decayed by 0.1 after every 10 epochs.

\section{Results and Discussion}
The results obtained using different strategies (as explained in Section 3) for training the I3D~\cite{carreira2017quo} model are shown in Table\ref{tab:result_val} and Table\ref{tab:result_test} for the validation and test dataset respectively. To our surprise, the I3D model trained using only RGB images performed the best on the test dataset, achieving an F1@5\% score of 66.05, even outperforming the I3D model trained using both RGB images and optical flow, which achieved an F1@5\% score of 65.95 on the test dataset. The I3D model, when used as a fixed feature extractor, obtained an F1@5\% score of 63.23 on the test dataset. On the validation set, both the I3D model trained on \textit{RGB} only, and the one trained on \textit{RGB with optical flow} gave results comparable to the baseline score of 52.11 \cite{shou2021generic}. In all our experiments we used the annotations corresponding to the annotator with highest F1-consistency score as groundtruth boundaries. In another approach we sampled the groundtruth annotation for every video based on the weights of the F1-consistency score of different annotators to train our models but couldn't achieve any performance boost with it. 

In order to interpret our results, we carried out a class based analysis of boundaries detected by our model. Figs.~\ref{fig:sfiga} and \ref{fig:sfigb} highlight 10 classes in KineticsGEBD with the highest and lowest mean success for correctly detected boundaries by our model respectively. We believe that classes like gymnastics\_tumbling and folding\_clothes have well understood the definition of boundaries and hence our model detects them consistently, whereas classes like using\_computer and shooting\_goal(soccer) have no precise and closed vocabulary for boundaries, making it difficult for our model to detect them. This is clearly illustrated in Fig.~\ref{fig:results_boundary_detection}.

An important point to observe is that we are using the Farneback algorithm to calculate  optical flow as it is computationally efficient. However, it is less accurate, which could be a possible explanation for the inferior performance. Furthermore our model does not explicitly learn changes in brightness or changes in camera angle or shots, which also justifies lack of performance in such cases. Lastly, Rel.Dis measures the discrepancy between the detected timestamp and the ground truth timestamp (if event boundary is a range then it is represented using the middle timestamp). However in research problems like hard cut shot boundary detection and gradual transition shot boundary detection, predictions within a window of time around the groundtruth timestamp are considered a correct prediction. We suggest incorporating such adjustments to the evaluation protocol would be more insightful and interpretable. An even more intricate version of this task would be to detect generic boundaries along with identifying semantic implications associated with them.

\section{Acknowledgement}
This work has emanated from research supported by Science Foundation Ireland (SFI) under Grant Number SFI/12/RC/2289\_P2, co-funded by the European Regional Development Fund and Xperi FotoNation.

{\small
\bibliographystyle{ieee_fullname}
\bibliography{sample}
}

\end{document}